\documentclass[letterpaper, 10 pt, conference]{ieeeconf}  
\usepackage{comment}
\usepackage{booktabs}
\usepackage{tabu}
\usepackage{graphicx}
\usepackage{caption}
\usepackage{lipsum}
\usepackage{cite}
\usepackage{amssymb}
\usepackage{pifont}
\usepackage{capt-of,etoolbox}
\newcommand{\ssymbol}[1]{^{\@fnsymbol{#1}}}
\usepackage[pagebackref=true,breaklinks=true,letterpaper=true,colorlinks,bookmarks=false]{hyperref}
\usepackage{adjustbox}
\usepackage{tablefootnote}
\usepackage{multirow}

\newcolumntype{N}{>{\centering\arraybackslash}m{0.55cm}}

\IEEEoverridecommandlockouts                              

\title{\LARGE \bf
CueCAn: Cue-driven Contextual Attention for Identifying Missing Traffic Signs on Unconstrained Roads
}
\author{Varun Gupta$^{\rceil}$, Anbumani Subramanian$^{\dag}$, C.V. Jawahar$^{\dag}$, Rohit Saluja$^{\ddag}$\\
\url{https://github.com/iHubData-Mobility/public-CueCAn}
\thanks{$^{\rceil \dag \ddag}$~The authors are with Center for Visual Information Technology (CVIT) Lab, IIIT Hyderabad, and IIT Mandi, India.
        {\tt\scriptsize $^{\rceil}$ varungupta.iiith@gmail.com},
        {\tt\scriptsize $^{\dag}$ \{anbumani,jawahar\}@iiit.ac.in},  {\tt\scriptsize $^{\ddag}$rohit@iitmandi.ac.in}}%
}

\begin{document}
\makeatletter
\let\@oldmaketitle\@maketitle
\renewcommand{\@maketitle}{\@oldmaketitle
  
  \includegraphics[width=\linewidth]
    {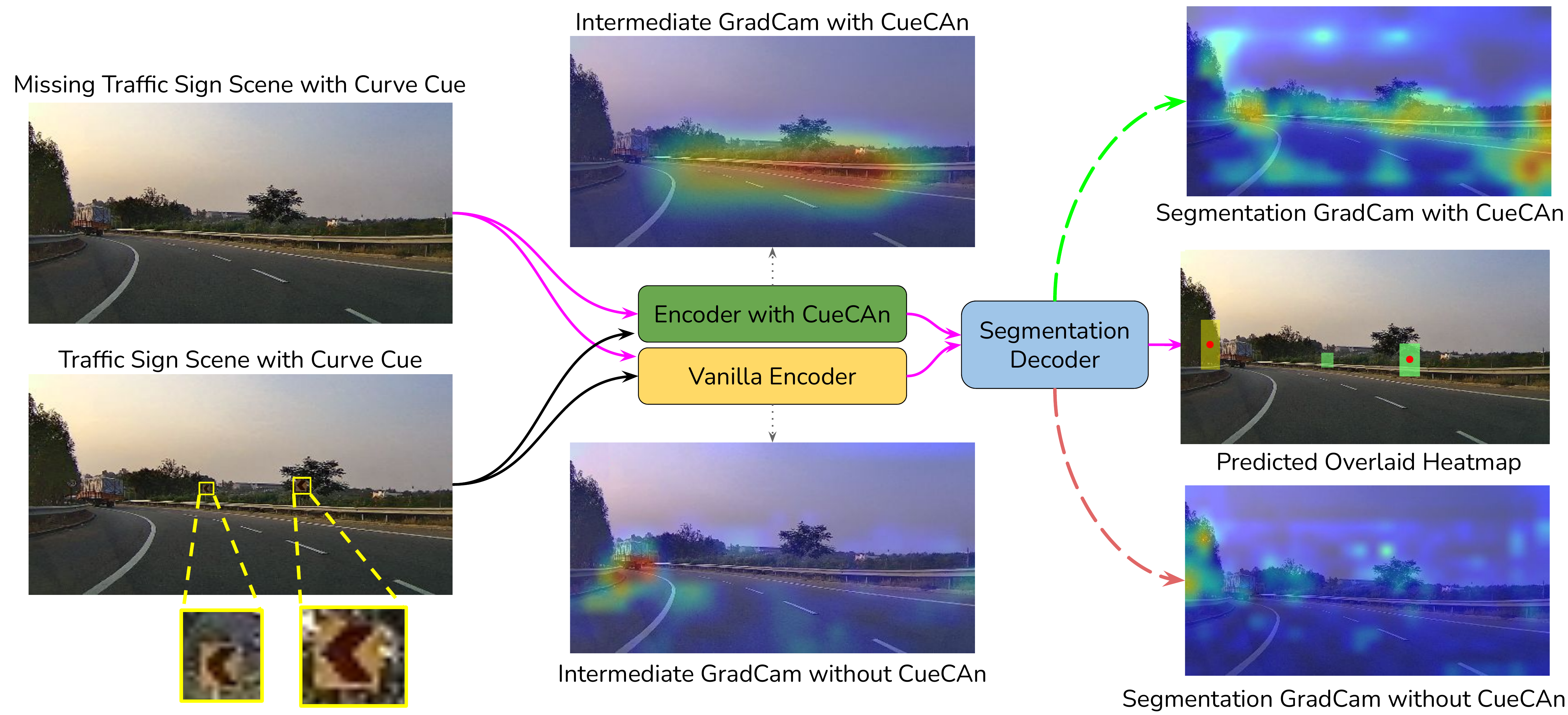}
    \captionof{figure}{{\bf Left:} Scenes with real and inpainted traffic signs (chevron-left). {\bf Middle:} Intermediary GradCAM visualizations of the cue classifier (encoder) with and without CueCAn. {\bf Right:} Segmentation model with CueCAn-based encoder detects missing signs (green masks overlayed over the scene on the right for CueCAn and yellow mask by the baseline) on the scene without signs (follow pink arrows) by effectively attending to the context cues, compared to weak attention without CueCAn. Segementation GradCAMs are obtained from the centroid of the predicted sign (red dot).}
    \label{fig:icra-banner}
    }
\makeatother
\maketitle
\setcounter{figure}{1}
\thispagestyle{empty}
\pagestyle{empty}


\begin{abstract}
Unconstrained Asian roads often involve poor infrastructure, affecting overall road safety. Missing traffic signs are a regular part of such roads. Missing or non-existing object detection has been studied for locating missing curbs and estimating reasonable regions for pedestrians on road scene images. Such methods involve analyzing task-specific single object cues. In this paper, we present the first and most challenging video dataset for missing objects, with multiple types of traffic signs for which the cues are visible without the signs in the scenes. We refer to it as the Missing Traffic Signs Video Dataset (MTSVD). MTSVD is challenging compared to the previous works in two aspects i) The traffic signs are generally not present in the vicinity of their cues, ii) The traffic signs' cues are diverse and unique. Also, MTSVD is the first publicly available missing object dataset. To train the models for identifying missing signs, we complement our dataset with 10K traffic sign tracks, with 40\% of the traffic signs having cues visible in the scenes. For identifying missing signs, we propose the Cue-driven Contextual Attention units (CueCAn), which we incorporate in our model's encoder. We first train the encoder to classify the presence of traffic sign cues and then train the entire segmentation model end-to-end to localize missing traffic signs. Quantitative and qualitative analysis shows that CueCAn significantly improves the performance of base models.
\end{abstract}


\begin{figure*}[t]
  \centering
  \includegraphics[width=1.0\linewidth]{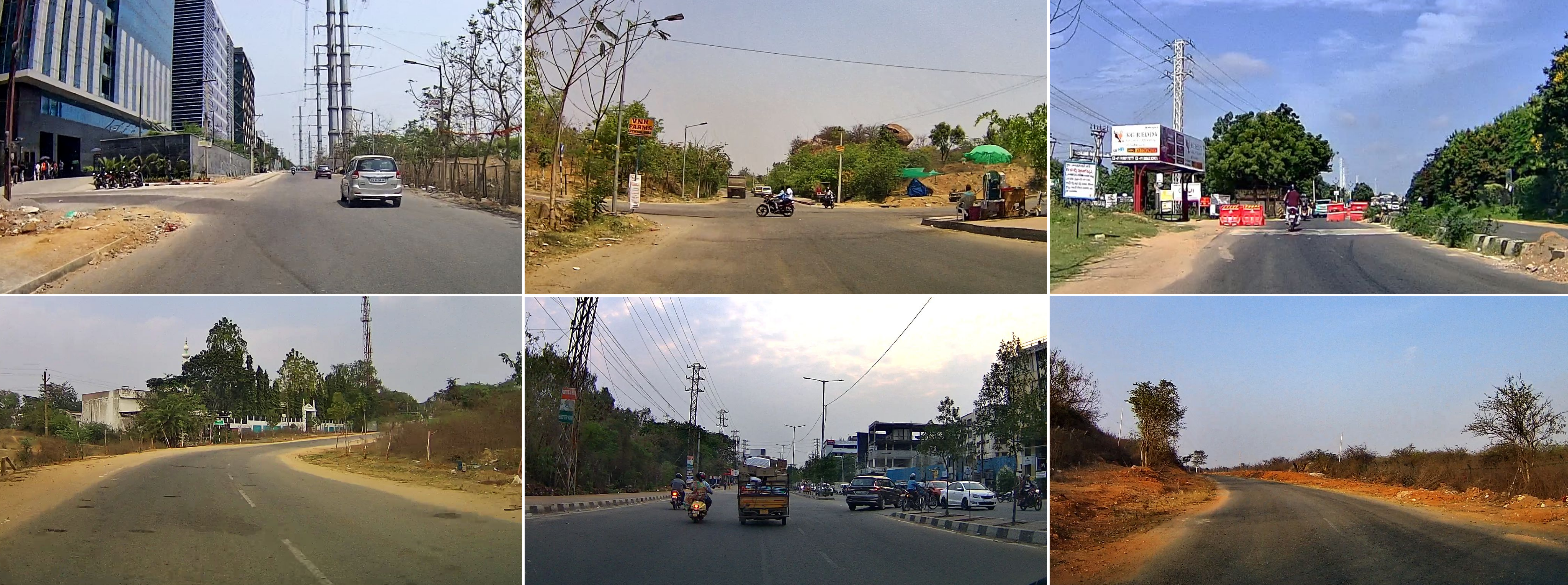}
  \caption{Sample scenes from MTSVD exhibiting missing traffic signs (with cues). \textbf{Top (left to right)}: side-road-left, cross-roads, go-slow and bus-bay. \textbf{Bottom (left to right)}: right-hand curve, gap-in-median and left-hand curve.}
  \label{fig:missing-samples}
\end{figure*}

\section{INTRODUCTION}
Road accidents affect millions of lives due to increasing population and vehicle density. Incorporating Advanced Driving Assistant Systems (ADAS) in commercial vehicles is gaining momentum, however, these systems still have a long way to go in terms of offering absolute road safety \cite{3decadesdriving}. In a recent road safety report, it is observed that $12.6\%$ of accidents caused by driver errors are due to traffic sign violations, which establishes the importance and need for traffic signs in a regulated manner \cite{AV_accidents}. Autonomous Vehicles (AV) require a consistent and well-maintained infrastructure to explore their full potential and deliver the safety they promise. Traffic signs are an important aspect of road infrastructure, as they contain essential information about what is coming ahead, which ADAS systems like Mobileye use as supplementary information for scene understanding. Failure to robustly perceive and process the road scene has led to multiple fatal crashes involving commercially deployed AVs as well \cite{traffic-sign-survey}. Sample frames from our Missing Traffic Signs Video Dataset (MTSVD), containing context cues but no signs are present in Fig. \ref{fig:missing-samples}. Identifying regions with missing traffic signs in a manual manner requires significant effort, and an automated solution to identify missing traffic signs regions will help authorities plan the infrastructure improvements required. Prior missing object datasets either had the cue in their immediate vicinity or had a consistent relationship between the missing object and the corresponding cue \cite{missingpedestrians,seeingwhatnotthere}. However, the MTSVD contains multiple cue contexts and complementary cue-object relationships, making MTSVD the most challenging and diverse, publicly accessible missing objects dataset. Based on the observation that context cues add a certain discontinuity to the scene, we propose CueCAn to exploit the nature of context cues. The CueCAn works on the intuition of identifying traffic sign cues by erasing/inpainting them (or filling the cue regions with context) in feature space and then taking the difference between inpainted and the original features, which helps in highlighting the discontinuous cue patterns. The proposed pipeline, as illustrated in Fig. \ref{fig:icra-banner}, highlights the efficacy of CueCAn in making the model attend to context features for both encoding and decoding tasks of classifying the presence of cues and localizing missing signs, which are lacking in the vanilla method. Our contributions are as follows:
\begin{itemize}
    \item We introduce the first publicly accessible video dataset for missing objects, the Missing Traffic Sign Video Dataset (MTSVD), spread across $10K$ video tracks for over $60$ traffic signs categories, with $2K$ missing sign video clips containing over $20$ types of traffic sign cues.
    \item We propose the Cue driven Context Attention Unit, \textit{CueCAn}, a cue-based approach for detecting missing traffic signs on unconstrained and traffic-dense roads.
\end{itemize}

\begin{table}
    \centering
    \caption{Overview of Missing Object Datasets.}
    \begin{tabu}{l c c c}
        \toprule
         \textbf{Dataset} & \textbf{Frames with} & \textbf{Public} & \textbf{Context}\\
         &\textbf{Missing Objects} & \textbf{Access}& \textbf{Variety}\\
         \midrule
        Tohme \cite{tohme, seeingwhatnotthere} & 1086  & $\times$ &$\times$ \\\midrule
         Pedestrians Dataset\cite{missingpedestrians} & $<$475 & $\times$ &$\times$ \\\midrule
         Missing Barricades \cite{missingbarricades} & 853  & $\times$ &$\times$ \\\midrule
         \textbf{MTSVD (ours)} & {\bf 135K (2K Videos)}  & {\bf \checkmark} &{\bf \checkmark} \\\bottomrule
    \end{tabu}
    \label{tab:missing-datasets}
    \vspace{-5mm}
\end{table}

\section{Related Work}
{\bf Missing Object Localization and Datasets}: Deep learning has driven computer vision approaches to be reasonably accurate for the tasks such as object detection and image segmentation. However, the task of locating missing objects is studied in very few works, but the existing ones highlight its potential \cite{seeingwhatnotthere,missingbarricades,missingpedestrians}. Humans consider multiple aspects to determine missing objects, the most prominent of which is cue understanding \cite{role-of-context,visual-object-incontext,context-in-wild}. Sun et al. \cite{seeingwhatnotthere} aim to improve city accessibility for people with mobility disabilities by detecting regions with missing curbs. For this, a Siamese Fully Connected network (SFC) learns the contextual-cue classification for curbs from the image and masked-image pairs, along with a curb localizer. The model identifies the region as a missing curb if an image contains curb context but no curb. Chian et al. \cite{missingbarricades} with an attempt to mitigate fall-from-height injuries, annotate the missing barricade regions, and perform object detection. Chien et al. \cite{missingpedestrians} predict the regions where pedestrians could be placed in street scenes using the Fully Convolutional Network (FCN) based on the VGG encoder. Grabner et al. \cite{grabner2010tracking} processed context in video tracking tasks using supporters, which helped estimate the target object locations using the Hough Transform.

An overview of existing missing object datasets is given in Table~\ref{tab:missing-datasets}. Given the lack of a large-scale and complex dataset specifically created for missing objects, prior missing object detection works either use existing standard datasets or collect and sample their own datasets lacking in scale and variety. Sun et al. \cite{seeingwhatnotthere} employ the TOHME dataset, a collection of $1086$ street ramp images sourced partly from Google Street View and crowd-sourcing \cite{tohme}. Chien et al. \cite{missingpedestrians} use the CityScapes dataset \cite{cityscapes}. However, more than $84\%$ of the CityScapes images contain fewer than $5\%$ pedestrian pixels, making most frames unsuitable for the task \cite{missingpedestrians}. Chian et al. \cite{missingbarricades} collected $853$ images of barricades captured from high-rise construction sites using a crane-mounted camera at varying elevations. However, these datasets do not involve diversity in the object cues. However, in the case of traffic signs, multiple cues exist, i.e., the cue for {\it left-hand-curve} is different from that of {\it right-hand-curve}, and for {\it bus-bay} is much different from that of {\it pedestrian-crossing}, etc.

\begin{table}[t]
    \addtolength{\tabcolsep}{-3.0pt}
    \centering
    \caption{Overview of Various Traffic Sign Datasets. \textit{Tracks} refer to the video tracks with signs in a sequence, and \textit{Missing signs} refer to intervals where cue-context exist, but the traffic sign does not. $\ddag$ created using video data. $\dagger$ 52K signs fully annotated and 48K partially annotated \cite{ertler2020mapillary}}.
    \begin{tabu}{l c c c c}
        \toprule
         \textbf{Dataset} & \textbf{Capture Resolution}  & \textbf{Frames} & \textbf{Night} & \textbf{Missing}\\
         & & \textbf{(Tracks)} & \textbf{+All-weather} & \textbf{Signs}\\
         \midrule
         DITS${^{\ddag}}$\cite{ItalianDITS} & 1280$\times$720 & $478$  & $\times$  & $\times$ \\\midrule
         TT100K\cite{chinatt100k} & 2048$\times$2048 & $26K$  & $\times$ & $\times$ \\\midrule
         GTSDS${^{\ddag}}$\cite{GermanTSDS} & 1360$\times$800 & $1206$  & $\times$ & $\times$ \\\midrule 
         BTSD\cite{belgiumTSDS-overview} & 1280$\times$720 & $8851$  & $\times$  & $\times$ \\
        \midrule
        MTSD\cite{ertler2020mapillary} & 2048$\times$1152  & $100K^{\dagger}$  &  \checkmark &  $\times$ \\\midrule
        MTSVD${^{\ddag}}$ & \textbf{2560$\times$1440} & {\bf 400K}  & \checkmark &  \checkmark \\
         (ours) & & {\bf (10K)} & &\\
         \bottomrule
    \end{tabu}
    \label{table:ts-datasets}
\end{table}
\begin{figure}[t!]
  \centering
  \includegraphics[width=\linewidth]{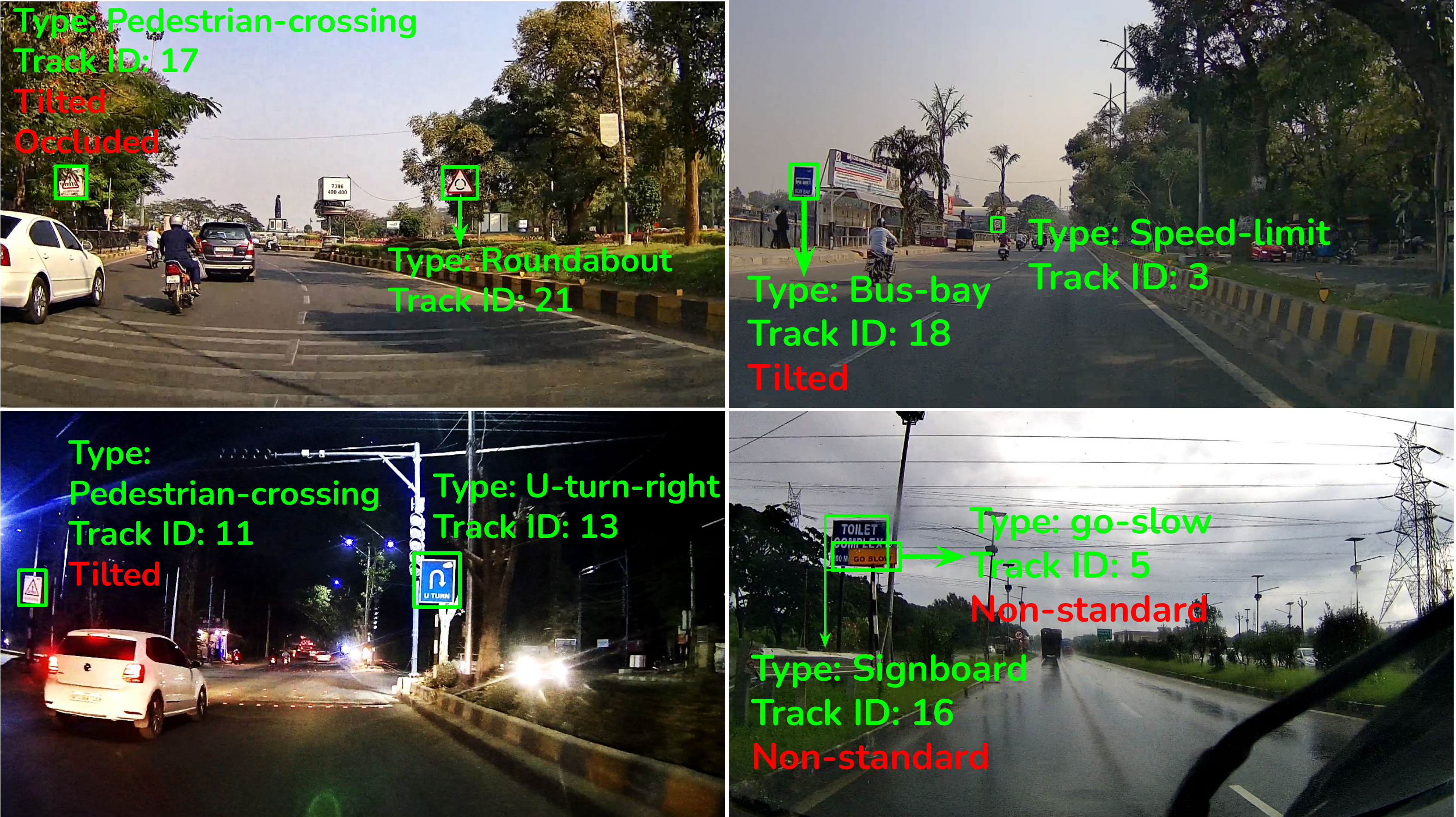}
  \caption{Annotated frames from MTSVD exhibiting frames having signs with their relevant cues. \textbf{Top} (left to right): a roundabout with pedestrian-crossing, bus-bay with speed-limit. \textbf{Bottom} (left to right): u-turn-right with pedestrian-crossing (night scene), go-slow and Signboard (rainy weather).}
  \label{fig:TS_and_Context}
  \vspace{-5mm}
\end{figure}
\begin{figure*}[t]
  \centering
  \includegraphics[width=0.9\linewidth]{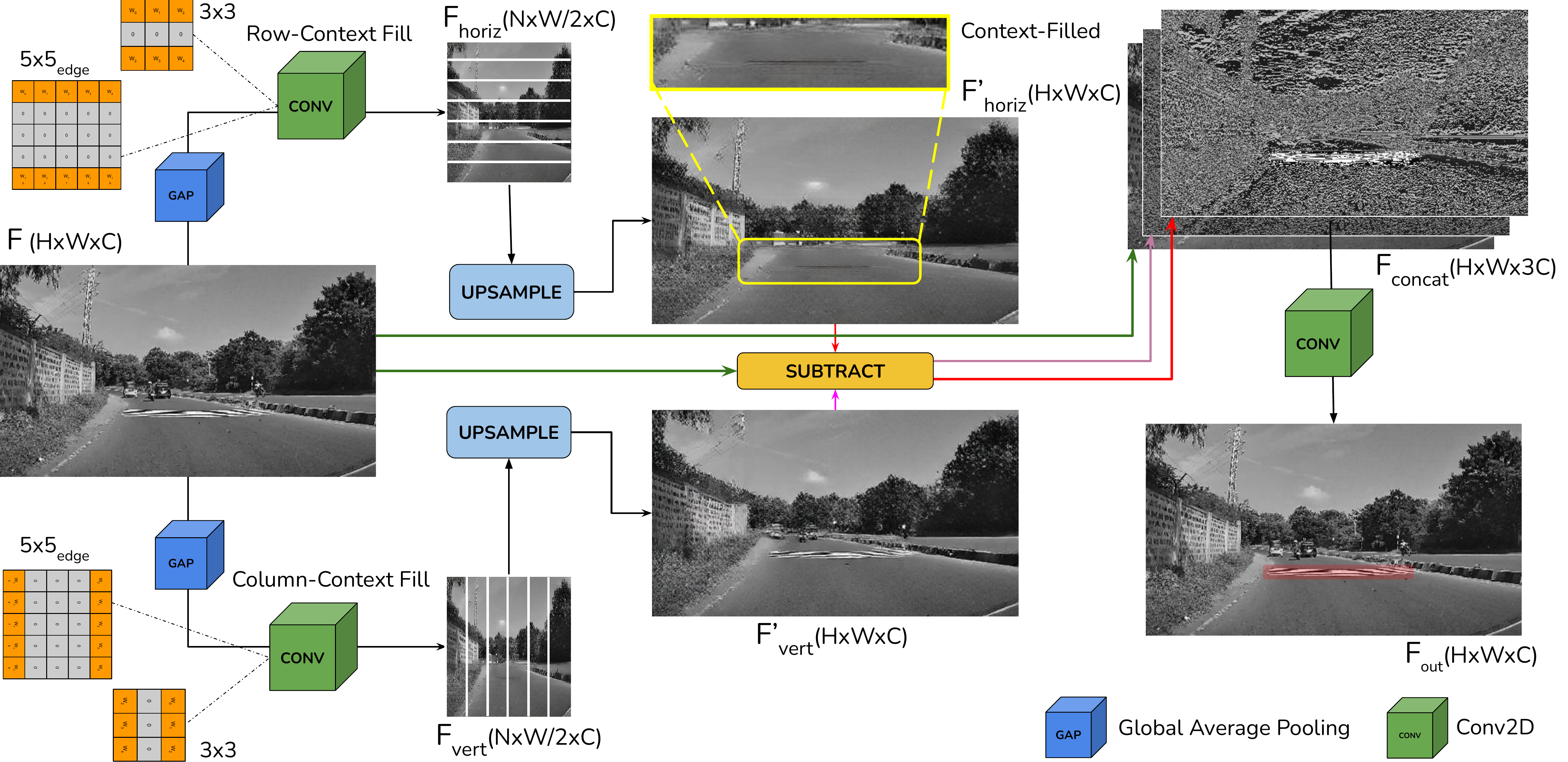}
  \caption{Illustration of CueCAn. Input feature map, F is downsampled using average pooling. Features are then inpainted using kernel sizes of 3 and 5, with different learnable units (gray: non-learnable, orange: learnable), in a row and column-wise fashion, and upsampled. Features are then subtracted from the original feature map and concatenated with the original feature. Finally, a Conv operation merges the features to obtain the original feature size.}
  \label{fig:cuecan}
  \vspace{-3mm}
\end{figure*}
{\bf Traffic Sign Datasets}:
Several traffic sign datasets exist \cite{ertler2020mapillary,ItalianDITS,chinatt100k,GermanTSDS,belgiumTS}. However, Traffic signs in different countries though sharing a similar context and purpose, have variations in appearance, promoting the compilation of regional datasets. Table \ref{table:ts-datasets} briefly summarizes available traffic sign datasets. Existing datasets focus on classification and segmentation tasks and leave out the missing signs data, resulting in the absence of a holistic approach toward road safety. The incentive behind MTSVD is to promote road safety on unconstrained roads, by enhancing the road infrastructure.

{\bf Context in Computer Vision}: Elaborate studies indicate how humans effectively perceive their surroundings via contextual clues \cite{role-of-context,visual-object-incontext,context-in-wild}. Santosh et al. \cite{santosh2009cvpr} elaborate on the role of context in human-scene perception, classify context into multiple categories and use context information to improve object detection. Multiple vision approaches integrate context and object features, like Conditional Random Fields, to improve object categorization and the Deformable Parts Model with surrounding context information to improve detection and segmentation tasks \cite{objects-in-context,context-in-wild}. However, given the close dependency between object features and their corresponding context, these are unsuitable for the current task, where the object remains absent from the scene.

{\bf Deep Neural Networks and Attention Mechanism}: Deep neural networks have gained immense popularity in multiple computer vision tasks. Fully Convolutional Neural (FCN) networks use skip connections to effectively capture contextual and spatial information from the features \cite{fcn}. FCN has been utilised for identifying missing pedestrians by Chien et al. \cite{missingpedestrians}, and the same has been used in this study for the segmentation task as well. For the segmentation tasks, the output matching the input size is generated with pixel-level class mappings. As we provide a plethora of data to neural networks, attention modules make the networks attend to specific input parts to improve the model’s performance. Initially, Bahdanau et al. \cite{bahdanau2014neural} introduced attention modules, which find applications in many AI applications, including Computer Vision, Natural Language Processing, Speech processing, etc. Liu et al. \cite{picannet} propose a contextual attention unit considering global and local object features for obtaining the saliency maps. However, local information in the spatial vicinity of the signs is irrelevant for identifying missing traffic signs, as they are predominantly at a location away from their cue. Further, global information is not essential for our task as the cues for each traffic sign are at specific locations in the scene. Han et al. \cite{RAU} provide a solution for detecting water puddles using the Reflection Attention Unit (RAU) on the basis of puddle surfaces containing reflection from the regions away from them (e.g., sky), which is similar to our situation (traffic signs away from their cue). However, RAU models search for similar corresponding patches in the continuous image space. In contrast, our task requires modeling traffic sign cues as a discontinuity in the image space, e.g., \textit{speed-breakers} (see Fig.~\ref{fig:cuecan}).

\section{Missing Traffic Sign Video Dataset}~\label{sec:missingvideodata}
In this paper, we present a novel dataset, MTSVD. What differentiates the MTSVD from other traffic sign datasets is the inclusion of $2K$ missing sign clips having traffic sign cues (refer to Table~\ref{tab:missing-datasets}). Further, it is the first publicly accessible data for missing objects. The dataset contains $135K$ frames of missing traffic sign intervals spread across $2K$ video clips covering over $20$ traffic sign classes for identifying, classifying, and detecting missing objects in images and video clips (refer to GitHub page for more information). The data is collected in Asia, covering a wide variety of terrain, weather, lighting conditions, and nighttime scenes. MTSVD is captured with the DDPAI X2S Pro camera, recording the driving at $25$ fps, with a resolution of $2560\times1440$. The MTSVD contains $10K$ unique traffic sign tracks, annotated with multiple attributes like occlusion, tilt, damage, and truncation. It also involves labels identifying non-standard traffic signs, like background-foreground color, shape, category, etc. Sample annotated images are shown in Fig~\ref{fig:TS_and_Context}, along with distinctive attributes in red. MTSVD is spread over $60$ categories of traffic signs.\footnote{The only other traffic sign dataset to include such dense attributes (6) and categories (313) for each annotation is the MTSD \cite{ertler2020mapillary} but does not enable missing object-related tasks.} Note that for safety purposes, sometimes the traffic signs appear before the cues in a video sequence (e.g., \textit{u-turn}). So, we also label the exact frame intervals when the traffic sign cues are in the camera's field of view.

\section{Methodology}\label{sec:method}
Similar to previous work on non-existent pedestrians \cite{missingpedestrians}, we employ VGG-19 \cite{vgg19} encoder followed by the FCN-8 decoder \cite{fcn} to segment the missing traffic sign regions. We augment VGG-19 layers with novel Cue-driven Contextual Attention (CueCAn) units and train the encoder to classify the presence or absence of traffic signs' cues, and further train the entire network with the enriched encoder features end-to-end for the segmentation task. We now discuss the components of our system in detail.
\subsection{Introducing CueCAn}
We model the context space of traffic sign cues as regions with discontinuities. E.g., in the case of \textit{speed-breakers} or \textit{go-slow} stripes on the road, this discontinuity is in the form of horizontal ridges. At the same time, the cues for \textit{gap-in-median} or \textit{side-roads} (see Fig~\ref{fig:missing-samples}) are complex discontinuities. We model such complexities by a composition of horizontal and vertical discontinuities in feature space. We propose highlighting these aberrations to focus on traffic signs' cues using CueCAn. The CueCAn learns to \textit{fill} the rows and columns of the image features with their context and finds the difference from the original feature vector. Non-filled regions with cues in the original features have a larger difference than their filled counterparts, which helps highlight the cues. Context cues without a linear geometry, like in the case of curves, benefit from the composition of horizontal and vertical filling. The implementation of the CueCAn is illustrated in Fig.~\ref{fig:cuecan}. First, the input feature map $F$ with shape $[H,W,C]$ is fed to an average pooling layer, condensing the feature to the shape of $[N,W/2,C]$. Here, $H$, $W$, and $C$ are the input feature's height, width, and channel dimensions, and N is fixed to $8$, similar to Han et al. \cite{RAU}. The features are then filled with the context in a row-wise fashion with a learnable $3\times3$ convolution operation with the central row fixed to zero (or depending on the encoder's layer: $5\times5$ convolution with central three rows fixed to zero), resulting in ${F_{horiz}}$ of shape $[N,W/2,C]$. A similar process is applied in a column-wise manner to fill the feature columns with context using convolutional filters with central columns fixed to zero, resulting in ${F_{vert}}$ of shape $[N,W/2,C]$. ${F_{horiz}}$ and ${F_{vert}}$ are upsampled (using bilinear interpolation) to the dimensions of original features, i.e. $[H,W,C]$, denoted by ${F_{horiz}^{'}}$ and ${F_{vert}^{'}}$, respectively. Finally, the upsampled features, now containing the context encoding, are individually subtracted from the input feature $F$, and are further concatenated with $F$, resulting in $F_{concat}$ of shape $[H,W,3C]$. The $F_{concat}$ is matched to the input feature $[H,W,C]$ using a Conv$+$ReLU operation. When filled with their context, regions such as the sky and road do not introduce a significant change. However, cue regions change more with the context filling operation, maximizing the difference between the original and the context-encoded feature (red highlight at the bottom right in Fig. \ref{fig:cuecan}), which provides supplementary information to the original feature vector by highlighting the traffic signs' cues.

\subsection{Training Data Creation}
To train the encoder for cue classification, we randomly sample $12500$ unique frames from MTSVD using the annotated intervals and tracks to create four balanced sub-sets. The sub-sets include i) frames containing traffic signs with cues, ii)  frames with traffic sign cues and inpainted traffic signs, using a state-of-the-art inpainting technique \cite{LAMA}, iii) frames without any cue but with traffic signs (from categories having no cues in the same frame, e.g., \textit{school-ahead}), and iv) frames without any cue or traffic sign. Thus we balance the dataset to avoid any bias toward the traffic signs or cues. The traffic sign cue is a central part of identifying missing traffic signs, so we only use the inpainted images and corresponding inpainting masks\footnote{as the ground truth for missing traffic signs doesn't exist.}  as input-output pairs for the localization (segmentation) task, similar to previous works \cite{seeingwhatnotthere,missingpedestrians}. It is important to note that despite using inpainted images, locating missing traffic signs is more challenging than previous works because i) MTSVD contains over $20$ complex traffic sign cues, 
ii) locating traffic signs is challenging even after attending to the cue regions due to the variety of possible traffic sign placements.

\subsection{Model Training}
To identify missing signs, the model must attend to the traffic sign cues in the environment. Similar to Han et al. \cite{RAU}, we add CueCAn at the end of the third, fourth, and fifth blocks of the VGG-19 \cite{vgg19} encoder to highlight and classify the cues. 
The next task is to localize \textit{where} the sign could be placed using the segmentation model with the pre-trained VGG-19 encoder and FCN-8 \cite{fcn} decoder. For localization, optimal results and GradCAM visualizations are observed when the entire network is fine-tuned end-to-end. We use the binary cross-entropy loss for classification and focal loss \cite{focal-loss} to handle class imbalance in the localization task.

\begin{figure*}[t]
  \centering
  \includegraphics[width=0.9\linewidth]{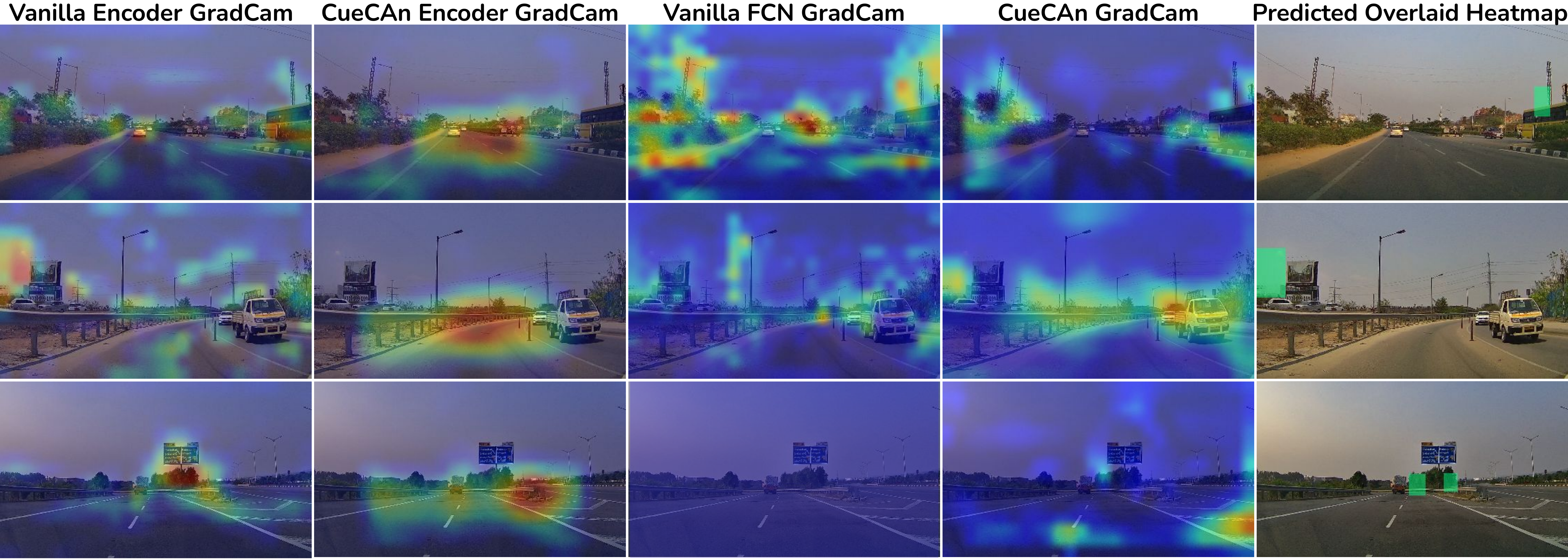}
  \caption{Illustration for localization of missing traffic signs. \textbf{Top}: cue for gap-in-median is attended well by the CueCAn. Overlaid predicted missing sign segmentation map (green mask) is placed rightfully at the gap's neck. \textbf{Mid}: cue for a right-hand-curve. The CueCAn kernels and feature compositions help attend to curves. The predicted segmentation map is rightly placed on the left side. \textbf{Bottom}: cue for gap-in-median, CueCAn places two signs, one very close to the junction and the other a bit farther away. Previous works fail to localize any sign for all three samples.} 
  \label{fig:result-matrix}
  \vspace{-5mm}
\end{figure*}
\section{Experiments and Results}~\label{sec:experiments}
We use a batch size of $32$ and train the cue classifier with \textit{Adam} optimizer with an initial learning rate of $1e^{-4}$. The segmentation model is also trained with \textit{Adam} optimizer and an initial learning rate of ${1e^{-3}}$. All models are trained on a single Nvidia RTX-2080Ti GPU for 400 training epochs, with train:val:test split of 80:10:10.

We consider the vanilla VGG-19 encoder to be the baseline cue classifier. We also experiment with two different versions of the CueCAn, i) by changing the convolutional kernel size and ii) by changing the learnable parameters of the convolution filter. We try convolutional kernel sizes of $3$, $5$, and $7$, with two context-filling approaches for each kernel size. We start with an implementation in which all kernel parameters except the central row/column are learnable. We refer to this configuration as CueCAn${_{k}}$ where $k$ is the kernel size. In the next version, CueCAn${_{ke}}$, only the boundary (or edge) rows or columns are learnable. The intuition behind the CueCAn${_{ke}}$ is that if only boundary features are used, it leads to better context filling in the feature space. Using only the edge parameters reduces the noise of nearby pixels (or elements in feature space) containing cues due to the receptive field. We verify the intuition mentioned above by the comparative GradCAM \cite{gradcam} visualizations using different configurations, with all other parameters kept constant.

Further, we implement arrangements of CueCAn at the end of all the blocks of the VGG-19 encoder. However, the training loss and accuracy curves indicated the ineffectiveness of this approach. We, thus, implement the final set of experiments with CueCAn in only the $3{^{rd}}, 4{^{th}}$, and the $5{^{th}}$ blocks of the network, similar to Han et al. \cite{RAU}. We refer to configurations with convolutional kernels of size $k$, $k'$, and $k''$ in blocks $3$, $4$, and $5$ as CueCAn$_{kk'k''}$. Similarly, CueCAn$_{kekek'}$ means that the configurations are similar to CueCAn$_{kkk'}$, but the kernels in blocks $3-4$ are \textit{edge-filling} convolutions. We first experiment with CueCAn$_{333}$ as a baseline. Motivated by the increasing receptive field in deeper layers, we experiment with CueCAn$_{553}$, CueCAn$_{753}$ to learn to fill more context in the cue regions. However, we observe that due to the receptive field, the contextual regions around the cues also have cue information, making it complex for the non-edge kernels to fill the cues. Hence, we also experiment with CueCAn$_{5e53}$ and CueCAn$_{5e5e3}$. As we will see in the next section, using kernels above $5$ degrades both precision and recall, we, therefore, avoided using kernels of dimension $7$ in the final configurations.

\begin{figure}[ht]
  \centering
  \includegraphics[width=\linewidth]{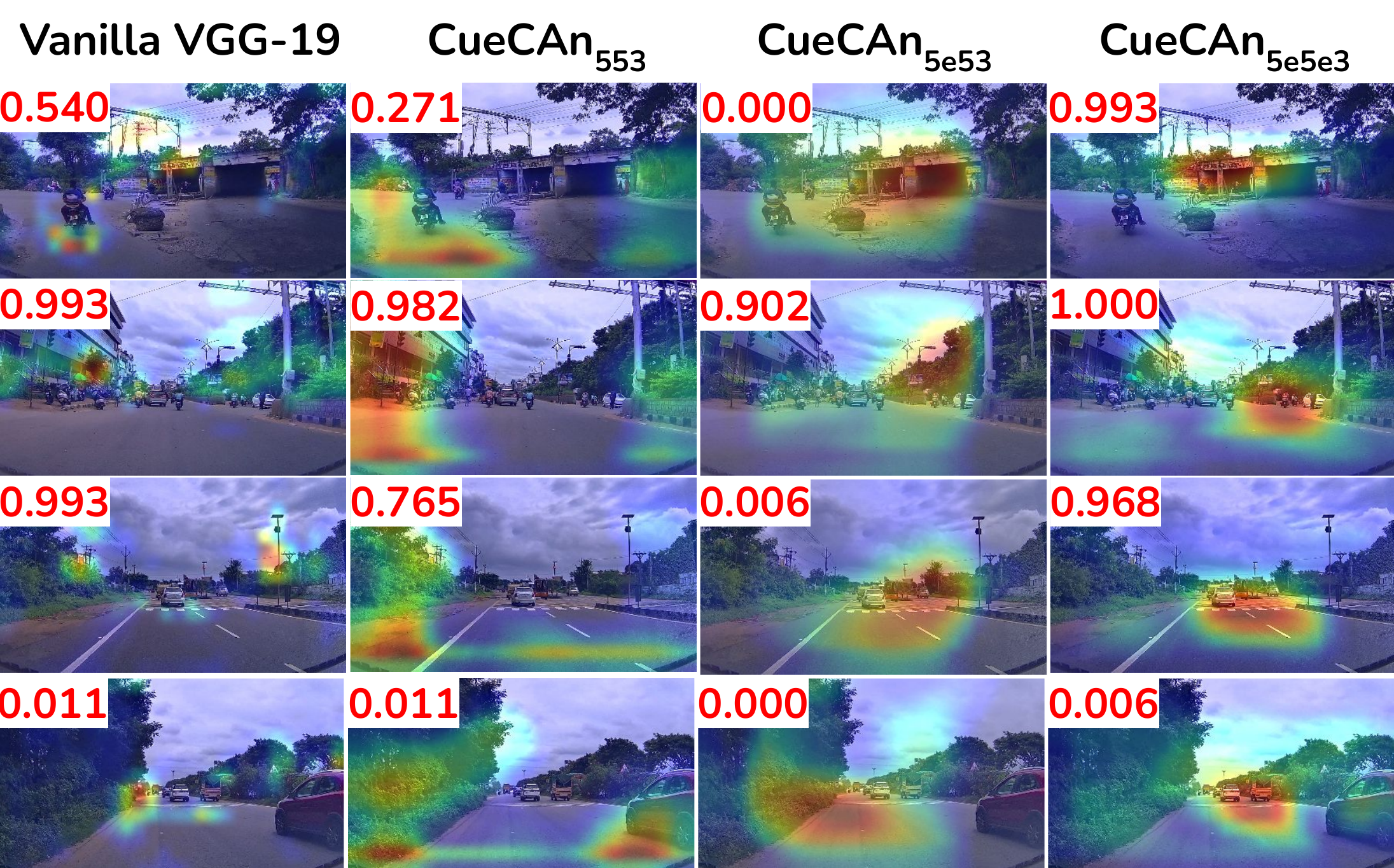}
  \caption{\textbf{Top-to-Bottom}: cues for {\it height-limit}, {\it gap-in-median}, {\it pedestrian-crossing}, and a failure case. The baseline fails to attend to context cues. CueCAn${_{553}}$ though attends to cues much better than the baseline, is observed to have low precision towards the important road regions.}
  \label{fig:result-ablation}
  \vspace{-5mm}
\end{figure}

\begin{table}[t]
    \centering
    \caption{Traffic Sign Cue Classification Results.}
    \begin{tabu}{l c c c}
    \toprule
         \textbf{Model} &  \textbf{Precision} & \textbf{Recall} & \textbf{F-Score} \\ 
         \midrule
         VGG19 \cite{vgg19} & 94.77 & 87.45  & 90.96 \\\midrule
         CueCAn${_{333}}$ & 94.87 & 93.96 & 94.41 \\\midrule
         CueCAn${_{553}}$ & 95.30 & 92.20 & 93.72 \\\midrule
         CueCAn${_{753}}$ & 92.48 & 90.99 & 91.73 \\\midrule
         CueCAn${_{5e53}}$ & 96.05 & \textbf{94.49} & 95.26 \\\midrule CueCAn${_{5e5e3}}$ & \textbf{97.96} & 93.22 & \textbf{95.53} \\\bottomrule
    \end{tabu}
    \label{tab:context-classifier-metrics}
    \vspace{-5mm}
\end{table}
{\bf Classification Results}: We present the traffic sign cue classification results in Table~\ref{tab:context-classifier-metrics}. It can be observed that the baseline VGG19 has precision, recall, and F-score of $94.77$, $87.45$, and $90.96$. Using VGG19 with CueCAn$_{333}$ shows significant improvements of over $3.5\%$ in recall and F-score. This is an impressive result since the model with CueCAn learns to better classify multiple types of traffic sign cues ($20$ categories as discussed in Sec.~\ref{sec:missingvideodata}) without any cue-level supervision. The classifier only uses a binary label (i.e., the presence or absence of a cue) at the frame level. Increasing the kernels' size of context filling filter from 3 to 5 in CueCAn$_{553}$ further improves the precision but reduces the recall, as shown in row $3$ of Table~\ref{tab:context-classifier-metrics}. The next row shows that increasing the kernel size to $7$ leads to the degradation of all three scores. However, using edge-kernels instead of kernels with only a central row fixed to zero (see Sec.~\ref{sec:method}) leads to further improvements in precision and F-scores as the last two rows of Table~\ref{tab:context-classifier-metrics} depict, though the recall of CueCAn$_{5e53}$ is better than CueCAn$_{5e5e3}$. Nevertheless, through qualitative analysis in Fig.~\ref{fig:result-ablation}, it can be observed that CueCAn$_{5e53}$, which is qualitatively also better than VGG-19 and CueCAn$_{333}$, fails to focus on the exact cue regions for {\it gap-in-the median} (second row). Moreover, the last column of Fig.~\ref{fig:result-ablation} shows that CueCAn$_{5e5e3}$ attends the correct traffic sign cues with high precision compared to the other variants. Thus, we use CueCAn$_{5e5e3}$ for the segmentation task and refer to it as CueCAn henceforth.

\begin{table}
    \addtolength{\tabcolsep}{-3.3pt}
    \centering
    \caption{Localization Results}
    \begin{tabu}{l c c c c c}
    \toprule
        {} & \textbf{Sun et al.\cite{seeingwhatnotthere}} & \textbf{Chien et al.\cite{missingpedestrians}} & \textbf{CueCAn} & \textbf{FCN-P} &  \textbf{CueCAn-P}\\\midrule
        Recall & {34.72} & {47.75} & {64.95} & {71.41} & {\bf 86.82} \\\bottomrule
    \end{tabu}
    \label{tab:segmentation-metrics}
    \vspace{-4mm}
\end{table}
{\bf Localization Results}: Table~\ref{tab:segmentation-metrics} lists the recall rate for different methods, as proposed by \cite{missingpedestrians,seeingwhatnotthere}, to localize non-existing pedestrians, and missing curbs. SFC \cite{seeingwhatnotthere} achieves the lowest recall of $34.72$ since it is designed to localize missing objects with cues around them. This only occurs for {\it obstacle-delineator} and {\it bus-bay} traffic signs in our data. Chien et al. \cite{missingpedestrians} (a variant that includes FCN) achieve a better recall rate of $47.75$. FCN, with the proposed CueCAn units, achieves the highest recall of $64.95$ compared to the previous approaches. We also post-process the predictions from the two localization models (FCN and CueCAn) by taking the tight rectangular region around the predicted blobs. As the last two columns of Table~\ref{tab:segmentation-metrics} depict, the two models with post-processing, i.e., FCN-P and CueCAn-P, significantly improve the recall rates, helping us achieve a recall of $86.82$. The result is impressive since GradCAM visualizations of the point at the center of predicted green rectangles in Fig.~\ref{fig:result-ablation} show that CueCAn attends to the correct traffic sign cues while simultaneously localizing the corresponding signs. We observe that the previous works fail to predict any mask for all Fig.~\ref{fig:result-ablation} samples (see the yellow mask in Fig.~\ref{fig:icra-banner} middle-right). The failure case for our approach is shown in the third row of Fig.~\ref{fig:result-ablation}, which predicts two traffic signs for a single cue, perhaps due to its distant location.
\begin{table}[t]
    \centering
    \caption{Results for Missing Traffic Sign Video Recognition}
    \begin{tabu}{l c c c}
    \toprule
         \textbf{Task} &  \textbf{Precision} & \textbf{Recall} & \textbf{F-Score}\\\toprule 
         Region classification & 59 & 60 & 59.49 \\ \midrule
         Video recognition & 37.50 & 50 & 42.85 \\\bottomrule
    \end{tabu}
    \label{tab:video-testing}
    \vspace{-5mm}
\end{table}

{\bf Results on Missing Traffic Sign Videos}: We use the CueCAn-P model for video recognition on $2K$ missing sign intervals in MTSVD. We also add $2K$ video clips without any missing sign cues to fairly test our model. We empirically observed that traffic sign predictions near the frame's central column are confusing cases as they represent cues that are far from the camera. Therefore, we use the predictions' centre, height, width, distance from the image center and aspect ratio, and train a Random Forest \cite{breiman2001random} to classify predicted regions into missing or non-missing (80:10:10 split). Finally, we use majority voting from all CueCAn-P's predictions in an interval for video recognition. Table~\ref{tab:video-testing} results show that missing traffic sign video recognition remains challenging.

\section{Conclusion and Future Work}
We presented the Missing Traffic Signs Video Dataset (MTSVD), a dataset for missing objects. Further, we propose a solution to identify missing signs using a CueCAn-based VGG-19 cue classification encoder coupled with the FCN-8 decoder to locate missing traffic signs. CueCAn fills the rows and columns of features with their context and subtracts the filled and the original features to highlight the traffic-sign cues. CueCAn significantly improves the results, qualitatively and quantitatively. In the future, we would like to explore MTSVD for multi-label missing sign detection, and propose this task as a time-series problem to investigate further the problems related to missing traffic signs.

{\bf Acknowledgement} The project is funded by the iHub-Data and Mobility at IIIT Hyderabad. We thank the data collection and annotation team for their effort.

{\small
\bibliographystyle{IEEEtran}
\bibliography{IEEEfull}
}


\end{document}